\documentclass[10pt,twocolumn,letterpaper]{article}

\usepackage{cvpr}
\usepackage{times}
\usepackage{epsfig}
\usepackage{graphicx}
\usepackage{amsmath}
\usepackage{amssymb}

\usepackage{rotating}
\usepackage{float}
\usepackage{adjustbox}
\usepackage{enumitem}
\newcommand{\id}{\mathrm{id}}
\setlist[1]{itemsep=-4pt, leftmargin=10pt}\usepackage{array}
\newcolumntype{P}[1]{>{\centering\arraybackslash}p{#1}}
\usepackage[pagebackref=true,breaklinks=true,letterpaper=true,colorlinks,bookmarks=false]{hyperref}

\cvprfinalcopy 

\def\cvprPaperID{****} 

\ifcvprfinal\fi
\begin{document}

\title{The Attack Generator: A Systematic Approach Towards Constructing Adversarial Attacks}

\author{Felix Assion$^1$, Peter Schlicht$^2$ \\ 
Florens Gre\ss ner$^1$, Wiebke G\"unther$^1$, Fabian H\"uger$^2$, Nico Schmidt$^2$, Umair Rasheed$^2$\\ 
$^1$neurocat GmbH, $^2$Volkswagen AG\\
{\tt\small fa@neurocat.ai, peter.schlicht@volkswagen.de}
}

\maketitle

\begin{abstract}
Most state-of-the-art machine learning (ML) classification systems are vulnerable to adversarial perturbations. As a consequence, adversarial robustness poses a significant challenge for the deployment of ML-based systems in safety- and security-critical environments like autonomous driving, disease detection or unmanned aerial vehicles. In the past years we have seen an impressive amount of publications presenting more and more new adversarial attacks. However, the attack research seems to be rather unstructured and new attacks often appear to be random selections from the unlimited set of possible adversarial attacks. With this publication, we present a structured analysis of the adversarial attack creation process. By detecting different building blocks of adversarial attacks, we outline the road to new sets of adversarial attacks. We call this the "attack generator". In the pursuit of this objective, we summarize and extend existing adversarial perturbation taxonomies. The resulting taxonomy is then linked to the application context of computer vision systems for autonomous vehicles, i.e.\ semantic segmentation and object detection. Finally, in order to prove the usefulness of the attack generator, we investigate existing semantic segmentation attacks with respect to the detected defining components of adversarial attacks.
\end{abstract}
\section{Introduction}

Recent advances in the field of machine learning have aroused the interest to apply these techniques in safety- and security-critical application contexts. One example is the integration of convolutional neural network-based dense classifiers into autonomous cars~\cite{Long, Gidaris}. In this challenging domain, we require not only a high accuracy on the true underlying data distribution, but also the trained ML module to be able to deal with maliciously crafted inputs. 

Unfortunately, the last few years have shown that current state-of-the-art ML algorithms, in particular deep neural networks, are quite brittle. With the publications of Szegedy \etal~\cite{Szegedy} and Goodfellow \etal~\cite{Good1} as a starting point, adversarial examples have been recognized as significant weak points. 

An adversarial example is an input data point that is slightly perturbed by an adversarial perturbation to cause misclassifications. These adversarial perturbations are created by an adversary with the help of an adversarial attack and are often hard to detect or even imperceptible to the human eye. The imperceptibility is not only challenging for the desired deployment in safety- and security-critical industries, but also hints at a crucial difference between the sensory information processing in humans and in artificial neural networks~\cite{Brendel}. Since the discovery of this vulnerability, a lot of different adversarial attacks and defenses have been published, \eg~\cite{Carlini2, Kurakin, Su}. It has become an arms race between attackers and defenders~\cite{Raghunathan}.

The development of new adversarial attacks remains to be one key objective of adversarial robustness research. This is due to the fact that adversarial attacks play a central role in the context of robustifying ML systems, as well as during the evaluation of adversarial robustness. 
For example, adversarial attacks are often part of a defense strategy. Currently, there does not exist any defense mechanism that is fully satisfactory, although adversarial training shows promising results. Adversarial training integrates adversarial examples into the training procedure, i.e.\ the neural network is trained on a mixture of clean and adversarial data points~\cite{Tramer, Kurakin2, Madry}. Thus, this defense strongly depends on adversarial attacks, which can provide the needed adversarial perturbations. 
At the same time, adversarial attacks are also central for the evaluation of whether or not a deep neural network is robust. Ideally, the robustness evaluation process should be independent of concrete attacks and instead, build on provable verification techniques, i.e.\ methods that can issue robustness guarantees~\cite{Wong, Dvijotham}. Unfortunately, these provable approaches are not yet scalable to complex tasks like semantic segmentation or object detection. As a consequence, one has to again rely on a set of adversarial attacks for the evaluation process. 

This raises the question how one can develop large, diverse sets of strong adversarial attacks, which can help with the hardening and the evaluation of neural networks. Although the attack research is flourishing, this question has not been answered. 
Even the leading software toolboxes, like the Adversarial Robustness Toolbox~\cite{Nicolae}, CleverHans~\cite{Papernot2} or the Foolbox~\cite{Rauber}, still offer rather limited collections of benchmark attacks. This also implies that most of the defense proposals are only evaluated against a handful of arbitrarily selected attacks. Furthermore, we still miss broadly accepted attack-based benchmark challenges for safety- and security-critical tasks. 
These constraints are the result of the current modus operandi in the development of new attacks. Adversarial attacks are basically published one by one with a fixed threat model in mind. For example, the first wave of adversarial attacks was very much fixated on white-box, targeted attacks with $L^p$-imperceptibility constraints for simple classification tasks~\cite{Yuan}. 

Up until now, we missed the chance to analyze adversarial attacks on a structural level. A structural analysis helps us understand the defining parts of an adversarial attack, i.e.\ see an adversarial attack as a composition of various elements. In this way, one could shift the research focus from arbitrarily assembled attacks to defining new potential elements of the detected building blocks of an attack. This view would directly increase the number of adversarial attacks significantly, since every new element implies a large number of new adversarial attacks, namely all potential combinations with other compatible building block elements. This modular structure of an attack has already been partially recognized by the research community within the discussion of imperceptibility metrics~\cite{Luo}. Every adversarial attack contains some kind of imperceptibility measure. Traditionally, there has been a strong focus on $L^p$-norms as the driver of imperceptibility to the human eye~\cite{Sharif}. Recently, a lot of publications suggest other quantifiers to measure perceptual similarity within an adversarial attack, \eg~\cite{Engstrom, Wasserstein}. This is already a significant progress, since every known adversarial attack can now be updated by exchanging $L^p$-balls with these new proposed measures. 

In this paper, we take a first step towards the detection of structural similarities between adversarial attacks. We acknowledge that adversarial attacks can be viewed as (constrained) optimization problems combined with optimization methods, which try to find a solution of the optimization problem. With the help of an adversarial perturbation taxonomy, we further define building blocks and various influencing factors of the optimization problem and optimization method of an adversarial attack. Finally, we test our conceptual ideas by analyzing prominent existing attacks. 

In summary, our key contributions are:
\begin{itemize} 
\setlength\itemsep{-0.5em}
\item We consolidate and extend existing adversarial perturbation taxonomy approaches. The different dimensions of the proposed taxonomy are then equipped with potential options for the adversary, which are loosely connected to the computer vision task for autonomous driving. However, the taxonomy can easily be applied to other domains by adjusting the options within the taxonomy dimensions. 
\item We argue that adversarial attacks are a composition of different quantifiers / measures, which can be grouped and can be directly linked to the different dimensions and options of the taxonomy. We then suggest a deeper investigation of new measures linked to the taxonomy dimensions. In this way, we pave the way to the fast generation of new attack sets, i.e.\ outline the "attack generator". 
\item We validate our conceptual ideas by investigating the semantic segmentation adversarial attacks introduced in~\cite{Metzen}. Furthermore, we present first small experiments, where we deduce new attacks by exchanging various measures of the original attack formulations.
\end{itemize}


\section{Taxonomy of Adversarial Perturbations}
In this section we want to taxonomize adversarial perturbations along multiple dimensions, hence describe different classes of adversarial perturbations. These classes are helpful in a variety of contexts. Especially when considering adversarial robustness as a security issue, it becomes crucial to analyze essential properties of a realistic threat. In the past, publications were largely concerned with perturbation classes, which do not relate to specific security concerns~\cite{Gilmer}. Thus, there is an obvious need to further clarify realistic threat scenarios, in order to close the gap between the literature and the concerns related to the actual deployment of ML systems. 
It has to be noted that what constitutes a relevant, realistic threat is highly application-specific. However, a general taxonomy can provide the necessary structural framework for this risk evaluation. 

Taxonomy approaches for adversarial perturbations, adversarial examples or adversarial attacks have already been presented in several publications, \eg~\cite{Serban, Carlini1, Yuan, Gilmer, Papernot1}. In the following, we consolidate and extend these taxonomy approaches. Additionally, we explore options within the different dimensions of the taxonomy. While the dimensions of the taxonomy are application independent, some of the options are motivated by the computer vision task for autonomous driving. The dimensions of this taxonomy proposal are inspired by the framework for empirical evaluation of classifier security presented in~\cite{Biggio, Biggio3}. 

In general, we recognize two central questions when classifying adversarial perturbations: Who created the adversarial perturbation? And which attack strategy led him to the perturbation at hand? As a consequence, the proposed taxonomy consists of the two high-level dimensions "threat model" and "attack strategy". The threat model summarizes the most important information about the adversary. Influenced by his goals, knowledge and constraints, the adversary then develops an attack strategy, which ultimately results in an adversarial attack and thus, the considered adversarial perturbation.

\subsection{Threat Model}

The threat model characterizes the attacker. It usually specifies his goals, knowledge and capabilities (constraints). Thus, we suggest to further decompose the threat model into these three sub-dimensions.

\subsubsection{Adversary's Goals}
\label{sec:goals}
The overall objective of the adversary is to force the victim model to make mistakes with the help of an adversarial perturbation. But this rather broad goal can be further specified by discussing the type of output the adversary desires (specificity) and defining the scope in which the perturbation should be successful in harming the ML system (perturbation scope). Furthermore, one key premise of an adversarial perturbation is that it should be imperceptible or inconspicuous. Since imperceptibility is still a very abstract concept, the adversary usually has a more specific type of imperceptibility in mind (perturbation imperceptibility). We will now go through the different aspects of the adversary's goals and equip them with suitable options for the adversary. It should be noted that options are not necessarily mutually exclusive. This will also be true for options presented in other dimensions of the taxonomy. \newline
\newline
\textbf{Specificity}: What are the desired consequences of the adversarial perturbation? 
\begin{itemize}
\item  \textit{Untargeted (Non-targeted)}: The goal is to craft a perturbation which results in as many misclassifications as possible. There is no preference concerning the appearing classes in the adversarial output~\cite{Arnab}. 
\item  \textit{Static Target}: The perturbation should lead to a fixed classification output, which is essentially independent of the input point added to the perturbation~\cite{Metzen}. For example, the perturbation always forces the victim model to output one fixed image of an empty street without any pedestrians or cars in sight. 
\item  \textit{Dynamic Target}: This type of goal has also been introduced by Metzen \etal~\cite{Metzen} in the context of attacking semantic image segmentation. Here, the adversarial perturbation aims at keeping the ML module's output unchanged with the exception of removing certain target classes. The desired classification output depends on the input point which is combined with the crafted perturbation. Removing the pedestrian class in every possible traffic situation is an example for a dynamic target objective.
\item  \textit{Confusing Target (Confusion)}: The adversarial perturbation should keep the classification output unchanged with the exception of changing the position or size of certain target classes. As in the dynamic target setting, the desired output is related to the considered input image. As an example, one can think of an adversarial perturbation that reduces the size of pedestrians and in this way leads to a false sense of distance. 
\end{itemize}
\textbf{Perturbation Scope}: What is the desired application scope of the adversarial perturbation? 
\begin{itemize}
\item  \textit{Individual Scope}: The perturbation is crafted for one specific input image, i.e.\ one specific
adversarial example is the target of the adversary. It is not necessary that the same perturbation fools the ML system on other data points. 
\item  \textit{Contextual Scope}: The goal is to create a fixed image-agnostic perturbation that causes label changes for one or more specific contextual situations. For example, the perturbation works for traffic situations on snowy or rainy days and is then able to fool the victim model under the majority of angles, distances and lighting effects.
\item  \textit{Universal Scope}: The goal is to create a fixed image-agnostic perturbation that causes label changes for a significant part of the true data distribution with no explicit contextual dependencies. This scope has first been proposed by Moosavi-Dezfooli \etal~\cite{Moosavi} and has been further analyzed in~\cite{Metzen, Perolat}.  
\end{itemize}
\textbf{Perturbation Imperceptibility}: In which way should the perturbation be imperceptible?
\begin{itemize}
\item  $L^p$\textit{-based Imperceptibility}: Due to small changes with respect to some $L^p$-norm, the human observer should not be able to detect the adversarial perturbation when applied to one or more input images.
\item  \textit{Attention-based Imperceptibility}: Due to unremarkable changes, the human observer should not be able to detect the adversarial perturbation when applied to one or more input data points. These unremarkable changes are not motivated by a $L^p$-norm, but are rather the result of other measures of perceptual similarity. Examples are perturbations based on rotations and translations~\cite{Xiao}, Wasserstein distance~\cite{Wasserstein} or SSIM~\cite{Sharif}.
\item  \textit{Output Imperceptibility}: A human observer can not easily detect irregularities in the classification output whenever the adversarial perturbation is applied. For instance, adversarial examples still lead to plausible traffic situations and misclassifications are integrated unobtrusively into their environment.
\item  \textit{Detector Imperceptibility}: A predefined selection of software-based detection systems is not able to detect irregularities in the input, output or in the activation patterns of the ML module caused by the adversarial perturbation. Hence, the adversary tries not only to mislead the victim model, but also adversarial example detectors placed around the victim model~\cite{Metzen, Meng}. 
\end{itemize}

\subsubsection{Adversary's Knowledge}
\label{sec:knowledge}
The knowledge of the adversary can be divided into "knowledge about the victim model and its parameters" and "knowledge about the training data set"~\cite{Biggio2}. The publications \cite{Kurakin3, Ilyas} were used as a basis for the following list of options.  \newline
\newline
\textbf{Model Knowledge}: What does the adversary know about the ML model and its parameters? 
\begin{itemize}
\item  \textit{White-box}: The adversary has full knowledge of the model internals, hence is aware of the concrete architecture, all parameter / weight configurations and possibly even the training strategy. 
\item  \textit{Output-transparent Black-box}: The adversary can not retrieve model parameters, but he can observe all or parts of the class probabilities or logits of the ML module's output.
\item  \textit{Query-limited Black-box}: The adversary can not access relevant model parameters, but he can observe the full or parts of the module's output on a limited number of inputs or with a limited frequency.
\item  \textit{Label-only Black-box}: The adversary can neither access relevant model parameters nor the class probabilities or logits, but he can observe the full or parts of the final classification decisions of the system, i.e.\ only access to inferred label (argmax layer).
\item  \textit{(Full) Black-box}: The adversary can neither retrieve relevant model parameters nor can he directly observe the output of the ML system. As a consequence, adversarial perturbations have to be created without querying the victim model.
\end{itemize}
\textbf{Data Knowledge}: What does the adversary know about the data sets which have been used to train the ML system?
\begin{itemize}
\item  \textit{Training Data}: The full or at least a significant part of the training data is available to the adversary.
\item  \textit{Surrogate Data}: There is no direct access to the original training data, but the adversary can collect data points from the relevant underlying data distribution of the victim model's environment. In the case of computer vision for autonomous driving, this is the minimal degree of data knowledge, since the adversary can always easily gather images or videos of traffic situations.
\end{itemize}

\subsubsection{Adversary's Capabilities}
\label{sec:cap}
Traditionally, this threat model characteristic clarifies the abilities and constraints of the adversary, thus outlines the attacker's power during his attempt to attack the ML system~\cite{Biggio2}. In this taxonomy we only investigate attackers utilizing adversarial perturbations. Thus, the capabilities of the adversary are fully defined by his means of feeding perturbations to the victim model. \newline
\newline
\textbf{Input Constraints}:  How can the adversary feed malicious input to the victim model?
\begin{itemize}
\item  \textit{Digital Data Feed (Direct Data Feed)}: The attacker can directly feed digital input to the ML module. Hence, he can adjust specific float values of input images.
\item  \textit{Physical Data Feed}: The adversary can not directly feed digital input, instead he creates physical perturbations, \eg~\cite{Athalye, Eykholt}. He has to place these adversarial objects in the environment of the autonomous car, which finally fool the module when they appear in the field of view of the camera. 
\item  \textit{Spatial Constraint}: It is not possible to place a physical or digital perturbation over the entire input image. Instead, the adversary can only influence limited areas of the input data.
\end{itemize}

\subsection{Attack Strategy}
\label{sec:strategy}
An adversarial perturbation is not fully characterized by the goals, knowledge and constraints of the adversary. One is still lacking a few fundamental decisions the adversary made on his way to the concrete formulation of the adversarial attack which in the end generated the perturbation. These decisions are always governed by the threat model. In other words, the taxonomy dimension "threat model" influences the decisions summarized in the "attack strategy". 

The attack strategy should specify what kind of model and data basis is going to be handed to the attack. Additionally, the structure of an adversarial perturbation differs strongly with the central mathematical procedure used within the adversarial attack to search for perturbation candidates. We therefore propose the following decomposition of the attack strategy.

\subsubsection{Attack Input}
\label{sec:inp}
With an adversarial perturbation the attacker wants to force the victim model to make classification mistakes. But, this does not imply that an adversarial attack is necessarily taking the true victim model into account during the generation of the perturbation. Analogously, the attacker has to decide what kind of data he wants to give to the attack and this can again deviate from the set defined by his data knowledge (see: Section \ref{sec:knowledge}). \newline
\newline
\textbf{Model Basis}: Which model is used by the adversarial attack? 
\begin{itemize}
\item  \textit{Victim Model}: The attack primarily utilizes the victim model in order to calculate adversarial perturbations.
\item  \textit{Surrogate Model}: The adversarial attack does not directly work with the victim model, but considers a surrogate model. This is often necessary if the adversary has only limited knowledge about the victim model or the victim model does not allow certain mathematical procedures~\cite{Liu}. 
\end{itemize}
\textbf{Data Basis}: Which data basis is used by the adversarial attack?
\begin{itemize}
\item  \textit{Training Data}: Data points of the victim model's original training data set are given to the adversarial attack.
\item  \textit{Surrogate Data}: The attack is primarily build on data that is related to the underlying data distribution of the task, but has not been previously used to train the ML system.
\item  \textit{No Data}: The adversary is not giving any task related data to the attack. Instead, the adversarial attack works with images that are not samples of the present data distribution~\cite{Du}.
\end{itemize}

\subsubsection{Mathematical Procedure}
With this dimension we try to summarize predominant mathematical tools that facilitate the detection of suitable adversarial perturbations. These tools are integrated into the adversarial attack itself. \newline
\newline
\textbf{Optimization Method}: Which mathematical procedure is the key ingredient for the perturbation search of the attack?
\begin{itemize}
\item  \textit{First-order Methods}: The adversarial attack tries to exploit perturbation directions given by exact or approximate (sub-)gradients.
\item  \textit{Second-order Methods}: The perturbation search is build on the calculation of the Hessian matrix or approximations of the Hessian matrix~\cite{Szegedy}.  
\item  \textit{Evolution \& Random Sampling}: The adversarial attack generates possible perturbations by sampling distributions and combining promising candidates. One can often fasten these methods by integrating prior knowledge about the decision boundary of the ML module~\cite{Brunner}. 
\end{itemize}

\section{The Attack Generator}
\label{sec: gen}
\begin{figure}[t]
\begin{center}
   \includegraphics[width=0.9\linewidth]{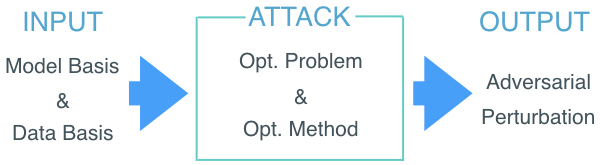}
\end{center}
   \caption{Adversarial attacks can be viewed as an optimization problem together with an optimization method. It takes some model and data set as input in order to create the perturbation.}
 \label{fig:1}
\end{figure}

An adversarial attack consists of two parts: (1) A constrained optimization problem that has to be minimized over admissible perturbations; (2) An optimization method that searches for approximate solutions of the constrained optimization problem. These two components of an attack are not always explicitly stated within an attack publication, but most of the time they are straightforward to derive. As input, the attack usually takes some kind of data set and a callable model. Potential choices with respect to the attack input have been discussed in the attack strategy dimension of the taxonomy (see: Section \ref{sec:inp}). On the other hand, the output of an adversarial attack is the desired adversarial perturbation or an adversarial example, i.e.\ a combination of the perturbation with a specific input data point (see: Figure \ref{fig:1}). 

Furthermore, the optimization method has also been introduced as a key part of the attack strategy. We presented various options of the adversary with "first-order methods" being the most common choice. Consequently, there is only one element of Figure \ref{fig:1} where we have not yet clarified its relation to the above presented taxonomy, namely the optimization problem of the attack.  
The optimization problem can abstractly be written as
\begin{equation}
\begin{split}
\label{eq:1}
\underset{\delta}{min} \;  [Obj(F,\mathcal{D})](\delta) \\
s.t. \;  \delta \in \mathcal{A},
\end{split}
\end{equation}
where $[Obj(F,\mathcal{D})](\cdot)$ is the objective function that takes a perturbation $\delta \in  \mathbb{R}^n$ as input and maps it to some fitness value in $\mathbb{R}$. Additionally, the objective function depends on the attack input and, in turn, on the provided ML-model $F$ and the data set $\mathcal{D}$. Often we can not take any arbitrary perturbation $\delta$, but we are rather constrained as introduced in the taxonomy dimension "input constraints" (see: Section \ref{sec:cap}). Thus, the given input constraints define an admissible set $\mathcal{A}$, which contains all potential perturbation candidates. 

Now, let us take a closer look at the objective function $[Obj(F,\mathcal{D})](\cdot)$: This function is the mathematical formalization of the goals of the adversary. For the adversary, minimizing the objective function is equivalent to achieving his goals with respect to specificity, perturbation imperceptibility and perturbation scope (see: Section \ref{sec:goals}). In order to arrive at this mathematical representation of his goals, the attacker has to initially define, directly or indirectly, quantifiers / measures that evaluate the level of specificity $\mathcal{M}_{sp}$, the level of imperceptibility $\mathcal{M}_{im}$ and the level of scope $\mathcal{M}_{sc}$. 
These are again real-valued functions which take the perturbation $\delta$ as input and additionally depend on the attack input, hence depend on $F$ and the full or parts of the provided data set $\mathcal{D}$. Thus, if one wants to be more thorough, one should rather write $[\mathcal{M}_{x}(F,\mathcal{D})](\delta)$ with $x\in \{sp,im,sc\}$. To make these abstract ideas a little bit more tangible, let us discuss a few examples for the different quantifiers which are frequently used in the adversarial attack literature: 

As already mentioned in the introduction, perturbation imperceptibility has in the past often been measured with the help of a $L^p$-norm, mostly $L^2$ or $L^{\infty}$. In these cases one has $\mathcal{M}_{im}(\cdot)= \| \cdot \|_{p}$. Please note that we in general do not pose any mathematical requirements on the real-valued maps $[\mathcal{M}_{x}(F,\mathcal{D})](\delta)$ with $x\in \{sp,im,sc\}$. If the adversary defines $\mathcal{M}_{im}(\cdot)= \| \cdot \|_{p}$, then $\mathcal{M}_{im}(\cdot)$ is a norm. But, we can also imagine situations where one might want to consider distance measures or imperceptibility quantifiers that do not fulfill the metric or norm axioms. For instance, if the adversary is interested in detector imperceptibility (see: Section \ref{sec:goals}), then imperceptibility of a perturbation is equivalent to a set of detectors not recognizing the attack. This imperceptibility measure does not follow the norm axioms, \eg due to binary output, measure is not absolutely homogeneous. 
In general, it is of utmost importance that the attack research looses its strong focus on $L^p$-norms as imperceptibility measures, since one can not expect that an adversary will do the favor of sticking to this one option of the perturbation imperceptibility taxonomy dimension.  

As a specificity measure $\mathcal{M}_{sp}$, attack researchers often make use of the original loss function $l(\cdot, \cdot)$ of the ML model. They insert the desired adversarial outcome $y^x_{tar}$ instead of the true label of data point $x \in \mathcal{D}$ and define $[ \mathcal{M}_{sp}(F,x)] (\delta) := l(F(x+\delta), y^x_{tar})$. In this example we see the usual dependence of $\mathcal{M}_{sp}(\cdot)$ on the input model $F$ and the input data set $\mathcal{D}$.

In the majority of existing attacks, the perturbation scope quantifier $\mathcal{M}_{sc}$ is closely connected to $\mathcal{M}_{sp}$. As discussed in the taxonomy, the desired scope defines in which situations $\mathcal{S} \subseteq \mathcal{D}$ the adversarial perturbation should be successful in harming the ML system (see: Section \ref{sec:goals}). To evaluate this, the adversary often takes Monte Carlo estimates over $\mathcal{M}_{sp}(\delta)$, thus
\begin{equation}
\label{eq:2}
\mathcal{M}_{sc}(\delta):= \frac{1}{N} \sum_{i=1}^{N} [ \mathcal{M}_{sp}(F,x_i)] (\delta),
\end{equation}
with desired scope data set $\mathcal{S} = \{x_1,...,x_N \}$ and $N$ being the cardinality of $\mathcal{S}$. 

\begin{figure}[t]
\begin{center}
   \includegraphics[width=0.92\linewidth]{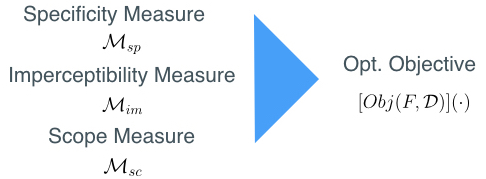}
\end{center}
   \caption{The optimization objective of an adversarial attack can be viewed as a composition of three adversary's goal measures.}
\label{fig:2}
\end{figure}

Finally, if one has determined these three goal measures, the objective function $[Obj(F,\mathcal{D})](\cdot)$ is just a composition of  $\mathcal{M}_{sp}$,  $\mathcal{M}_{im}$ and $\mathcal{M}_{sc}$. In other words, the selection of these three measures essentially defines the optimization problem of the adversarial attack (see: Figure \ref{fig:2}).
Going back to our previous examples, a sample composition is 
\begin{equation}
\begin{split}
\label{eq:3}
[Obj(F,\mathcal{D})](\delta) &:=  \mathcal{M}_{sc}(\delta) + \gamma \mathcal{M}_{im}(\delta) \\
&= \frac{1}{N} \sum_{i=1}^{N}  [ \mathcal{M}_{sp}(F,x_i)] (\delta)  + \gamma \| \delta \|_{p} \\
&= \frac{1}{N} \sum_{i=1}^{N}  l(F(x_i+\delta), y^{x_i}_{tar})  + \gamma \| \delta \|_{p},
\raisetag{5em}
\end{split}
\end{equation}
where $\gamma \in \mathbb{R}_{>0}$ is a weighting factor. This gives us the following attack optimization problem
\begin{equation}
\begin{split}
\label{eq:4}
\underset{\delta}{min}  \;  \frac{1}{N} \sum_{i=1}^{N}  l(F(x_i+\delta), y^{x_i}_{tar})  + \gamma \| \delta \|_{p} \\
s.t. \;  \delta \in \mathcal{A}.
\end{split}
\end{equation}
A lot of the published attack optimization problems introduce $\mathcal{M}_{im}(\delta)$ as an additional constraint instead of penalizing it in the objective function. In the setting of our example, this would lead to the following optimization problem
\begin{equation}
\begin{split}
\label{eq:5}
\underset{\delta}{min}  \;  \frac{1}{N} \sum_{i=1}^{N}  l(F(x_i+\delta), y^{x_i}_{tar})   \\
s.t. \;  \delta \in \mathcal{A}, \;  \| \delta \|_{p} \leq \epsilon
\end{split}
\end{equation}
with $\epsilon >0$ imperceptibility constant. With an appropriate choice of the weighting constant $\gamma$, Equation (\ref{eq:4}) and (\ref{eq:5}) lead to similar, or sometimes even the same, solutions and therefore, this does not significantly undermine our perspective on the attack problem presented in Equation (\ref{eq:1}). 

Overall, this gives us the insight that an adversarial attack consists of various building blocks, which are all linked to dimensions and options of the adversarial perturbation taxonomy (see: Appendix \ref{sec:A}). Creating a new adversarial attack is now equivalent to assembling adversary's goal measures to form an optimization objective and equipping this with a suitable optimization method. The choice of the optimization method has to acknowledge constraints given by the input model and input data as well as additional constraints on the perturbation. 

This modular view on an adversarial attack also outlines the path to the creation of sets of adversarial attacks instead of publishing one attack at a time. We have seen that the specificity, imperceptibility and scope quantifiers crucially define the adversarial attack. Thus, by investigating new measures of these kinds, one implicitly provides a number of new adversarial objective functions, namely all possible combinations with other adversary's goal measures. Finally, this results in a set of new adversarial attacks. 
In the context of computer vision systems for autonomous driving, researchers could therefore go through the options listed in Section \ref{sec:goals} and assign suitable quantifiers. This approach also helps us derive new adversarial attacks from existing ones by inserting alternative adversary's goal measures. In the following, we will underline the benefit of our conceptual ideas by experimenting with the attacks presented in~\cite{Metzen}.  

\section{Experiments}
\begin{figure*}[t]
\begin{center}
   \includegraphics[width=0.9 \linewidth]{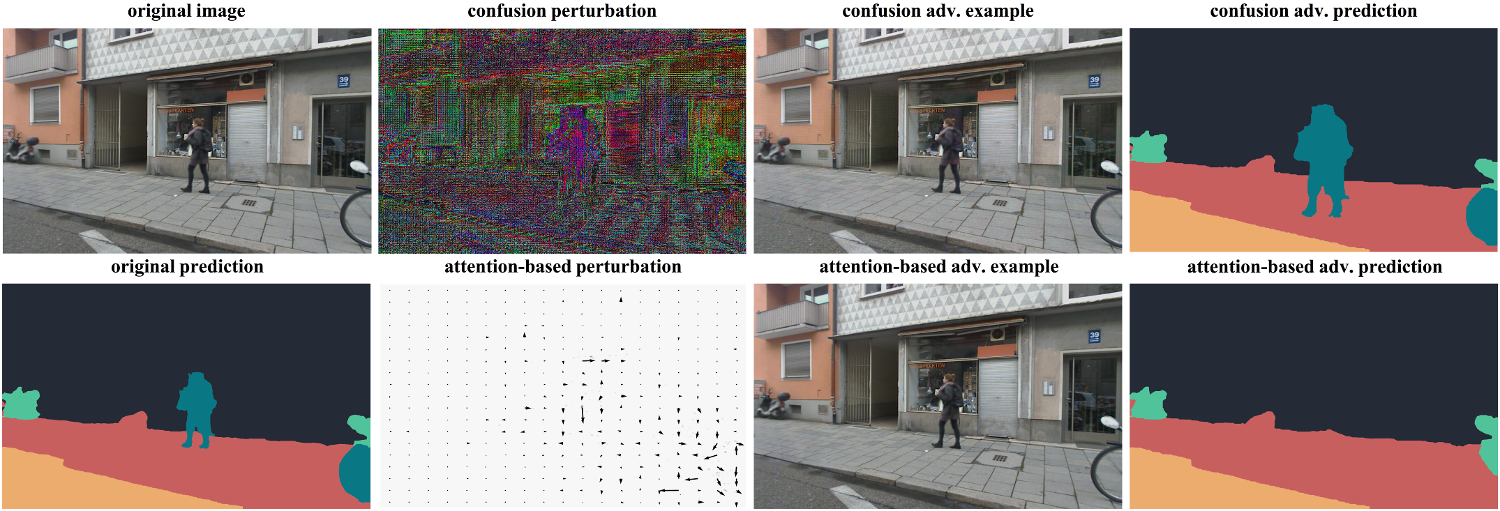}
\end{center}
   \caption{Sample results of adapted semantic segmentation attacks on ICNet with single image as attack input ($N$=1): (1) First column: Attack input image with original prediction of ICNet; (2) First row of right three columns - confusion: Attack enlarges pedestrian class ($\epsilon$ = 15); (3) Second row of right three columns - attention-based imperceptibility: Attack removes pedestrian class with flow field perturbation.}
 \label{fig:3}
\end{figure*}
We want to analyze two attacks introduced in~\cite{Metzen} to further clarify the concepts presented in Section \ref{sec: gen}. Additionally, we show how the modular view facilitates the deduction of new adversarial attacks from existing ones. 

Metzen \etal~\cite{Metzen} showed the existence of targeted, universal adversarial perturbations for state-of-the-art semantic segmentation neural networks. 
To generate these perturbations, Metzen \etal try to solve 
\begin{equation}
\begin{split}
\label{eq:6}
\underset{\delta}{min}  \;  \frac{1}{N} \sum_{i=1}^{N}  l(F(x_i+\delta), y^{x_i}_{tar})   \\
s.t. \;  \delta \in \mathbb{R}^{m \times n \times 3}, \;  \| \delta \|_{\infty} \leq \epsilon,
\end{split}
\end{equation}
where $\mathcal{D}=\{x_1,...,x_N \} \subseteq \mathbb{R}^{m \times n \times 3}$ is the whole training data set of the victim model $F$. The model output $F(x)$ consists of class probability vectors for every pixel of the input image $x \in \mathbb{R}^{m \times n \times 3}$. Equivalently to the example of Section \ref{sec: gen}, the function $l(\cdot, \cdot)$ denotes the loss function of the ML module, i.e.\ in this semantic segmentation setting 
\begin{equation}
\label{eq:7}
l(F(x),y):= \frac{1}{| \mathcal{I} |} \sum_{(i,j) \in \mathcal{I}} \mathcal{J}_{cls} (F(x)_{i,j},y_{i,j}),
\end{equation}
with $(i,j) \in \mathcal{I}$ spatial dimensions of an image and $\mathcal{J}_{cls}(\cdot,\cdot)$ the cross entropy classification loss. 

To solve the optimization problem of Equation (\ref{eq:6}), they follow an iterative gradient descent scheme, thus they exploit the white-box knowledge of the victim model by using a first-order optimization method. However, the key contribution of Metzen \etal is the proposed generation of the adversarial targets $y^{x}_{tar}$. As already mentioned in Section \ref{sec:goals}, they distinguish between a static and dynamic specificity target. In the static target case, one specific target segmentation is chosen for all input images, i.e.\ $y^{x}_{tar}:=y_{st}$ for all $x \in \mathcal{D}$.
For the dynamic target of removing a certain classification class, $y^{x}_{tar}:=y^{x}_{dy}$ is determined by applying a nearest-neighbor heuristic to the predicted classification decision $y^{x}_{pred}$ of the network. To be more precise, one substitutes all one-hot vectors of the target class by one-hot vectors which encode the nearest alternative non-target class. 

Now, let us take a look at the static and dynamic attack with the attack generator perspective of Section \ref{sec: gen}: As the imperceptibility measure we clearly have $\mathcal{M}_{im}(\cdot)=  \| \cdot \|_{\infty}$, i.e.\ imperceptibility is measured by the $L^{\infty}$-norm of the perturbation. The two attacks differ in their specificity objective, namely 
\begin{equation}
\begin{split}
\label{eq:8}
\mathcal{M}^{st}_{sp}(\delta)=  l(F(x+\delta), y_{st}) \\
\mathcal{M}^{dy}_{sp}(\delta)=  l(F(x+\delta), y^{x}_{dy}),
\end{split}
\end{equation}
with adversarial targets $y_{st}$ and $y^{x}_{dy}$ generated as described above. The scope measure is identical for the static as well as for the dynamic attack formulation. It is just the Monte Carlo estimation of the chosen specificity measure over the whole training set, thus 
\begin{equation}
\label{eq:9}
\mathcal{M}_{sc}(\delta)= \frac{1}{N} \sum_{i=1}^{N}  [\mathcal{M}^{c}_{sp}(F,x_i)](\delta),
\end{equation}
with $c \in \{st,dy \}$. We recognize again that the attack optimization objective (see: Equation (\ref{eq:6})) is a composition of the just defined adversary's goal measures and we are in an analog setting as in the example of Section \ref{sec: gen} (see: Equations (\ref{eq:4}), (\ref{eq:5})). 

After having worked out the different building blocks of the attacks, one can now think about exchanging different elements in order to derive new attacks on semantic segmentation modules. Recall that an adversarial attack consists of an optimization problem and an optimization method (see: Figure \ref{fig:1}). Thus, one potential adaptation is the selection of a different optimization method. Within the attack strategy taxonomy dimension we provided two options other than first-order methods. Especially applying evolution \& random sampling strategies might be beneficial, because they facilitate a perturbation search even if the adversary does not have full knowledge about the semantic segmentation module. However, we want to focus on changes concerning the attack optimization problem given by Equation (\ref{eq:6}). Changes here are basically equivalent to exchanging one or more of the three adversary's goal measures. 

Metzen \etal discuss static and dynamic targets, but they do not address untargeted or confusion specificity objectives (see: Section \ref{sec:goals}). A potential confusion goal could be to enlarge a target class, \eg increase size of pedestrian class. This can be achieved by substituting the original specificity measures by the very similar measure $\mathcal{M}^{co}_{sp}(\delta)=  l(F(x+\delta), y^{x}_{co})$, where $y^{x}_{co}$ is the adversarial confusion target for input image $x \in \mathcal{D}$. The only difference is the generation of the target segmentation $y^{x}_{co}$. Inspired by the original versions of the attacks, we again use a nearest-neighbor heuristic together with the predicted classification decisions $y^{x}_{pred}$ to craft $y^{x}_{co}$. However, this time we exchange the one-hot vectors of the nearest-neighbors of our target class and always insert the one-hot vector of the target class. This automatically leads to a target segmentation with an enlarged target class. For the implementation of this target generation, we used the nearest-neighbor interpolation of the OpenCV resize method~\cite{OpenCV}. Keeping all other adversary's goal quantifiers the same, this gives us a confusion semantic segmentation attacks. Figure \ref{fig:3} shows a sample result of this adapted attack on a self-trained ICNet for real-time semantic segmentation~\cite{Zhao} (see also: Appendix \ref{sec:B}). 

If one wants to keep the initial static and dynamic specificity measures, we could alternatively experiment with different imperceptibility quantifiers $\mathcal{M}_{im}$. Within the proposed taxonomy, we presented attention-based imperceptibility as an alternative option to $L^p$-based imperceptibility. This imperceptibility option contains a lot of interesting perturbation concepts, \eg adversarial perturbations generated through spatial transformation~\cite{Xiao}. In the spatial transformation setting, the adversarial perturbation is a flow field $\delta \in \mathbb{R}^{m \times n \times 2}$ which summarizes the per-pixel transformations of an image $x$ in order to get to the adversarial example $x^{adv}$. Hence, $x^{adv}=[T(\delta)](x)$ with $T$ being the function that applies the transformations of $\delta$ to the original image $x$. 
As an imperceptibility measure, one can then consider the total variation $TV(\cdot)$ of the flow field $\delta$:
\begin{equation}
\begin{split}
\label{eq:11}
\mathcal{M}_{im}(\delta) &= TV(\delta) \\
&=\sum_{(i,j) \in \mathcal{I}} \; \sum_{(k,l) \in \mathcal{N}(i,j)} \| \delta_{(i,j)}-\delta_{(k,l)} \|^2,
\end{split}
\end{equation}
where $\mathcal{N}(i,j)$ are the image coordinates of the 4-pixel neighbors of coordinate $(i,j) \in \mathcal{I}$. Note that $\delta$ is a flow field and hence $\delta_{(i,j)} \in \mathbb{R}^2$ for any spatial dimension $(i,j) \in \mathcal{I}$. 
With this in mind, we can formulate an attention-based version of the presented semantic segmentation attacks:
\begin{equation}
\begin{split}
\label{eq:12}
\underset{\delta}{min}  \;  \frac{1}{N} \sum_{i=1}^{N}  l(F([T(\delta)](x_i)), y^{x_i}_{tar}) + \gamma \; TV(\delta)  \\
s.t. \;  \delta \in \mathbb{R}^{m \times n \times 2},
\end{split}
\end{equation}
with $\gamma$ weighting factor and $y^{x_i}_{tar}$ the static or dynamic target label of image $x_i$. In Figure \ref{fig:3}, we present an exemplary result of this attack on a traffic situation containing a pedestrian (see also: Appendix \ref{sec:B}).

\section{Conclusion}
In this paper, we present a comprehensive adversarial perturbation taxonomy together with options for the adversary linked to every taxonomy dimension. We describe adversarial attacks as a composition of various elements which are closely related to the options of the given taxonomy. In particular, we illustrate the crucial role of adversary's goal measures in the creation of new adversarial attacks. This structured view on adversarial attacks facilitates the construction of sets of new attacks by investigating new specificity, perturbation imperceptibility and perturbation scope measures. Our experimental adaptations of existing semantic segmentation attacks demonstrate the benefits of this modular view on adversarial attacks. 

We propose a change of the publication style of adversarial attacks. We are convinced that a stronger focus on the exploration of new potential attack building blocks, instead of presenting fully assembled attacks, will help structure the adversarial attack research field and furthermore, will fasten the development of large, diverse sets of benchmark adversarial attacks.

\newpage

{\small
\bibliographystyle{ieee}
\bibliography{egbib}
}
\cleardoublepage


\appendix
\begin{table}{\section{Analysis Existing Attacks}\label{sec:A}}
		\centering
    	\begin{adjustbox}{angle=90}
    	\begin{tabular}{ | P{1.9cm} | P{5.4cm} | P{2.9cm}  | P{3.1cm} | P{3.1cm} | P{2.8cm} |}
    \hline
    \textbf{Attack Name} & \textbf{Opt. Problem} & \textbf{Opt. Method} & \textbf{Specificity} \newline $\mathcal{M}_{sp}(\delta)$ & 			    \textbf{Scope} \newline $\mathcal{M}_{sc}(\delta)$ & \textbf{Imperceptibility} \newline $\mathcal{M}_{im}(\delta)$ \\ \hline
    
    FGSM~\cite{Good1} &
	\begin{equation*}    
    \begin{split}
	\underset{\delta}{min}  -l(F(x+\delta),y^x) \\
	s.t. \;  \| \delta \|_{\infty} \leq \epsilon,
	\end{split}
	\end{equation*}
  	with $y^x$ true label of image $x$ and $l(\cdot,\cdot)$ loss function of neural network. & 
  	\textit{First Order Method}\newline Use scaled sign of gradient. & 
	\textit{Untargeted}\newline Measure: $$-l(F(x+\delta),y^x)$$ & 
	\textit{Individual Scope}\newline Only considering single image. Measure: $$\id (\mathcal{M}_{sp}(\delta))$$ & 
	$L^p$\textit{-based Imperceptibility}\newline Measure: $$\| \delta \|_{\infty}$$ \\ \hline
    
    L-BFGS~\cite{Szegedy} & 
    \begin{equation*}    
    \begin{split}
	\underset{\delta}{min} \;  l(F(x+\delta),y^x_{tar}) + \| \delta \|_{2}\\
	s.t. \; x+\delta \in [0,1]^m
	\end{split}
	\end{equation*}
	with $y^x_{tar}$ adversarial target. & 
    \textit{Second Order Method} \newline Box-constrained L-BFGS is a quasi-Newton method.  & 
    \textit{Targeted}\newline $y^x_{tar}$ is desired outcome. Measure: $$ l(F(x+\delta),y^x_{tar})$$ &
    \textit{Individual Scope}\newline Only considering single image. Measure: $$\id (\mathcal{M}_{sp}(\delta))$$ &  
    $L^p$\textit{-based Imperceptibility}\newline Measure: $$\| \delta \|_{2} $$ \\ \hline
   
    PGD~\cite{Madry}  & 
    \begin{equation*}    
    \begin{split}
	\underset{\delta}{min}  - l(F(x+\delta),y^x) \\
	s.t. \;  \| \delta \|_{\infty} \leq \epsilon,
	\end{split}
	\end{equation*}
  	with $y^x$ true label of $x$ and $l(\cdot,\cdot)$ loss function of neural network. & 
  	\textit{First Order Method}\newline Multi-step gradient descent combined with projections. & 
  	\textit{Untargeted}\newline Measure: $$-l(F(x+\delta),y^x)$$ & 
 	\textit{Individual Scope}\newline Only considering single image. Measure: $$\id (\mathcal{M}_{sp}(\delta))$$ & 
	$L^p$\textit{-based Imperceptibility}\newline Measure: $$\| \delta \|_{\infty}$$ \\ \hline

    Boundary \newline Attack~\cite{Brendel} & 
     \begin{equation*}    
    \begin{split}
	\underset{\delta}{min}  \;  \| \delta \|^2_{2} \\
      s.t. \;  c(\delta)=1
	\end{split}
	\end{equation*} 
	with $c(\cdot)\in \{0,1\}$ adversarial criterion for data point $x$. & 
	\textit{Evolution \& Random Sampling}\newline Method is initialized from an adversarial point and then random walk along decision boundary. &
	\textit{Untargeted / Targeted}\newline Depends on choice of adversarial criterion (and initialization point). Measure: $$c(x+\delta)$$ & 
	\textit{Individual Scope}\newline Only considering single image. Measure $$\id (\mathcal{M}_{sp}(\delta))$$ &  
	$L^p$\textit{-based Imperceptibility}\newline Measure: $$\| \delta \|^2_{2}$$ \\ \hline
	
	 EOT~\cite{Athalye} & 
	 \begin{equation*}    
     \begin{split}
	 \underset{\delta}{min} - \mathbb{E}_{t \sim T}[logP(y^x_{tar} \mid t(x+\delta))]\\
	 s.t. \;  \mathbb{E}_{t \sim T}[d(t(x),t(x+\delta))] \leq \epsilon,\\ 
	  x \in [0,1]^m
	 \end{split}
	 \end{equation*}  
	 with distribution $T$ of transformation functions, distance function $d(\cdot,\cdot)$ and $y^x_{tar}$ target.  & 
	 \textit{First Order Method}\newline Use projected gradient descent. & 
	 \textit{Targeted} \newline Measure also takes transformation $t$ as input. Measure: $$ -logP(y^x_{tar} \mid t(x+\delta))$$ & 
	 \textit{Contextual} \newline Consider one image $x$ and relevant transformations of $x$. Measure: $$\mathbb{E}_{t \sim T}[\mathcal{M}_{sp}(t,\delta))]$$ & 
	$L^p$\textit{-based / Attention-based Imperceptibility}: Depends on choice of distance. Measure: $$d(\cdot,\cdot)$$ \\ \hline
    	\end{tabular}
  	\end{adjustbox}
  	  \label{Tab:1}
  	  \caption{Building block analysis of existing adversarial attacks.}
    \end{table}

\newpage

\onecolumn
\begin{figure}{\section{Sample Results}\label{sec:B}}
\begin{center}
  \includegraphics[width=1 \linewidth]{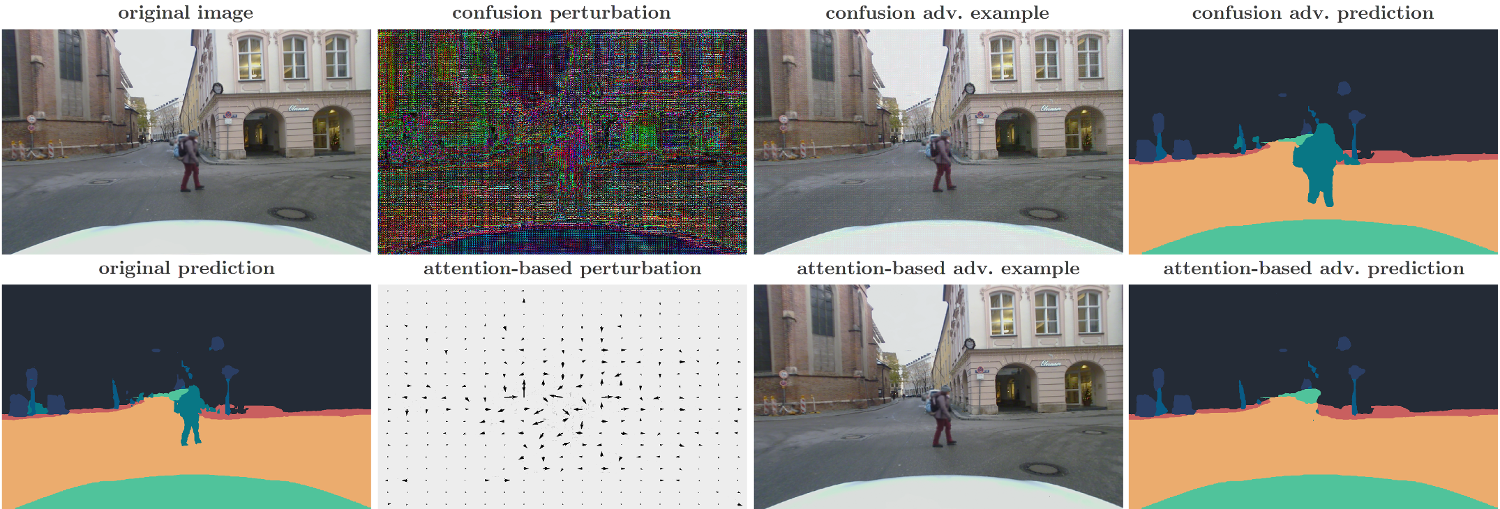} \\[2em]
  \includegraphics[width=1 \linewidth]{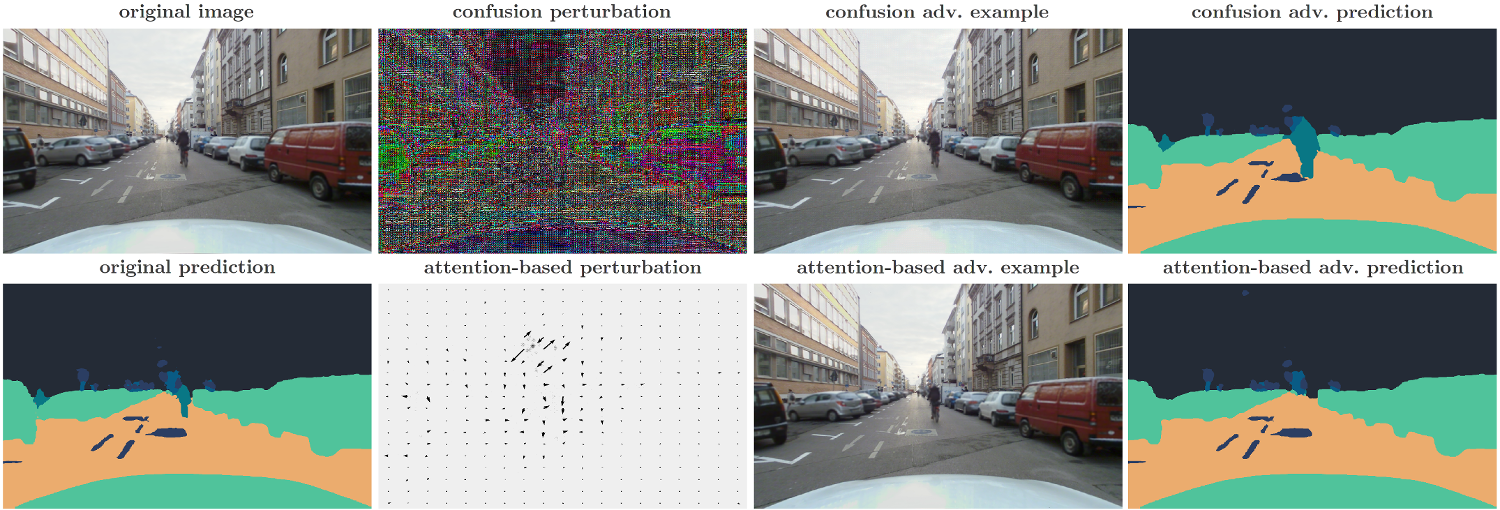} \\[2em]
  \includegraphics[width=1 \linewidth]{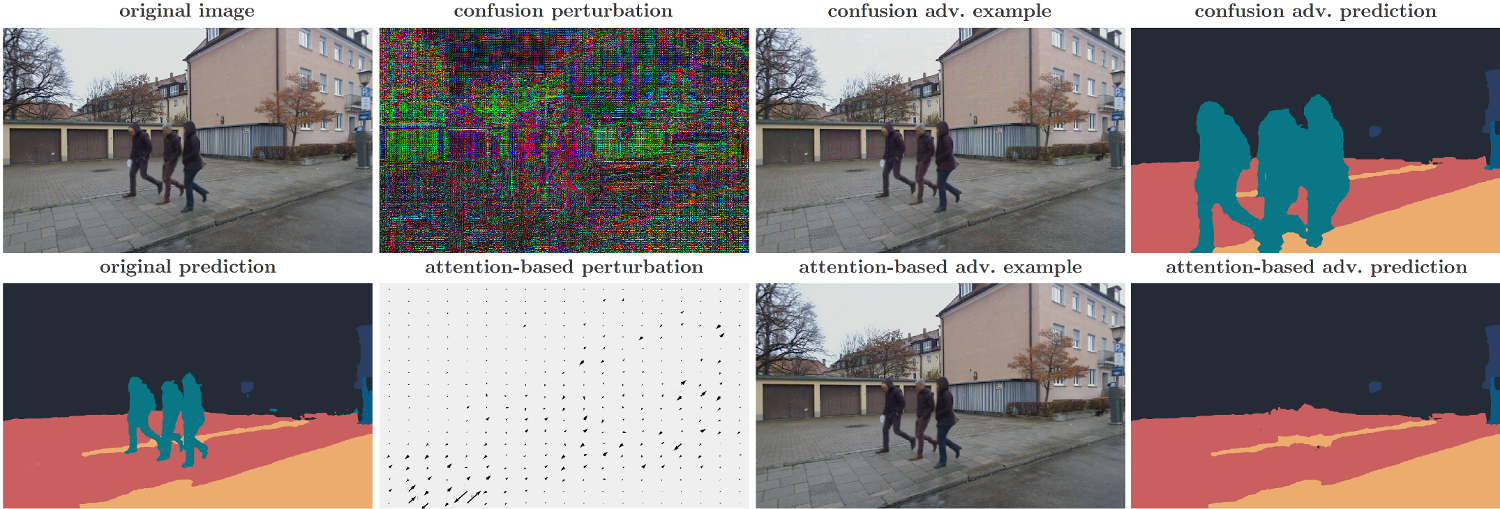} 
\end{center}
   \caption{Sample results of adapted semantic segmentation attacks with confusion target or attention-based imperceptibility. Attack input is a self-trained ICNet and a single image ($N$=1, $\epsilon$ = 15).}
 \label{fig:4}
\end{figure}

\end{document}